\documentclass{article}
\usepackage{arxiv}
\usepackage[utf8]{inputenc} 
\usepackage[T1]{fontenc}    
\usepackage[hidelinks]{hyperref} 
\usepackage{url}            
\usepackage{booktabs}       
\usepackage{amsfonts}       
\usepackage{nicefrac}       
\usepackage{microtype}      
\DeclareUnicodeCharacter{0301}{\'} 
\usepackage{amssymb}
\usepackage{lipsum}
\usepackage{graphicx}
\usepackage{float}
\usepackage{amsmath}
\usepackage[version=3]{mhchem}
\usepackage{soul}
\usepackage{color}
\usepackage{authblk}
\usepackage{threeparttable}
\usepackage{tabularx}
\graphicspath{ {./Figures/} }
\usepackage{xcolor}
\usepackage{xr}
\usepackage[normalem]{ulem}
\newcommand{\floridachem}{Department of Chemistry, University of Florida, Gainesville, FL 32611, USA}
\newcommand{\floridaqtp}{Quantum Theory Project, University of Florida, Gainesville, FL 32611, USA}
\newcommand{\floridamse}{Department of Materials Science \& Engineering, University of Florida, Gainesville, FL 32611, USA}
\newcommand{\floridaphys}{Department of Physics, University of Florida, Gainesville, FL 32611, USA}
\newcommand{\nyuphysics}{Center for Soft Matter Research, Department of Physics, New York University, New York 10003, USA}
\newcommand{\nyusimons}{Simons Center for Computational Physical Chemistry, Department of Chemistry, New York University, New York 10003, USA}
\newcommand{\nyucourant}{Courant Institute of Mathematical Sciences, New York University, New York 10003, USA}
\newcommand{\nyucns}{Center for Neural Science, New York University, New York 10003, USA}

\newcommand{\minnesotaaem}{Department of Aerospace Engineering and Mechanics, University of Minnesota, Minneapolis, MN 55455, USA}

\begin{document}
\title{MolGuidance: Advanced Guidance Strategies for Conditional Molecular Generation with Flow Matching}

\author[1,2$^\dagger$]{Jirui Jin }
\author[1,2$^\dagger$$*$]{Cheng Zeng}
\author[2,3]{Pawan Prakash}
\author[4]{Ellad B. Tadmor}
\author[1,2]{Adrian Roitberg}
\author[2,5]{Richard G. Hennig}
\author[6,7,8,9$*$]{Stefano Martiniani}
\author[1,2$*$]{Mingjie Liu}
\affil[$^\dagger$]{These authors contribute equally.}
\affil[1]{\floridachem}
\affil[2]{\floridaqtp}
\affil[3]{\floridaphys}
\affil[4]{\minnesotaaem}
\affil[5]{\floridamse}
\affil[6]{\nyuphysics}
\affil[7]{\nyusimons}
\affil[8]{\nyucourant}
\affil[9]{\nyucns}
\def\thefootnote{*}\footnotetext{c.zeng@ufl.edu, sm7683@nyu.edu, mingjieliu@ufl.edu}

\maketitle
\begin{abstract}
Key objectives in conditional molecular generation include ensuring chemical validity, aligning generated molecules with target properties, promoting structural diversity, and enabling efficient sampling for discovery. Recent advances in computer vision introduced a range of new guidance strategies for generative models, many of which can be adapted to support these goals. In this work, we integrate state-of-the-art guidance methods---including classifier-free guidance, autoguidance, and model guidance---in a leading molecule generation framework built on an SE(3)-equivariant flow matching process. We propose a hybrid guidance strategy that separately guides continuous and discrete molecular modalities---operating on velocity fields and predicted logits, respectively---while jointly optimizing their guidance scales via Bayesian optimization. Our implementation, benchmarked on the QM9 and QMe14S datasets, achieves new state-of-the-art performance in property alignment for \textit{de novo} molecular generation. The generated molecules also exhibit high structural validity. Furthermore, we systematically compare the strengths and limitations of various guidance methods, offering insights into their broader applicability.
\end{abstract}

\section*{Introduction\label{sec:Introduction}}
The generation of novel molecular structures with desired properties is crucial for drug design and chemical discovery~\cite{du2024machine, sanchez-lengeling2018}. Traditional molecular design using high-throughput screening and structure-property relationship-based rational design relies on testing or iteratively modifying known compounds, but it probes only a minute fraction of the $10^{60}$ chemical space and is slow, costly, and biased by human intuition.~\cite{polishchuk2013estimation, wouters2020estimated} In comparison, deep learning-based generative models have shown immense promise in narrowing down search space and accelerating molecule discovery. \cite{gomez-bombarelli2018, hoogeboom2022}
These models learn a probabilistic representation of the vast chemical space and then directly sample molecules with desired properties. 
A key challenge in this domain is to effectively guide the generation process towards molecules that are valid, novel, and in satisfactory alignment with property constraints. This process is also referred to as property-guided generation.

The most straightforward approach of property-guided generation is vanilla conditional generation, which takes the property as the condition and molecule representation (\textit{e.g.}, a molecular graph) as joint inputs, and learns a shared representation for properties and molecular structures that drives the sampling.~\cite{hoogeboom2022, xu2023} 
While vanilla conditional generation proves
simple and effective, it often struggles with precise property control—generated molecules may exhibit weak alignment
with target properties or deviate from desired values. To address these limitations, guidance techniques have been developed to bias the sampling trajectory toward outputs that more closely satisfy the desired conditions. The core principle underlying guidance methods is to enhance conditional generation by steering the model toward
property-specific molecular distributions. Various guidance methods have been proposed to achieve this goal.
Seminal guidance methods range from classifier guidance~\cite{dhariwal2021}, to classifier-free methods (CFG)~\cite{ho2022, zeni2025, nisonoff2025} which aim to improve generation quality with better property alignment. More recent approaches, such as autoguidance (AG)~\cite{karras2024} and model guidance (MG)~\cite{tang2025} further seek to balance property adherence with sample diversity and computational efficiency. 
Given the proliferation of these guidance strategies, a systematic comparison is crucial for researchers and practitioners to understand their relative strengths, weaknesses, and suitability for conditional molecule generation. 

A central challenge in molecular generation lies in developing effective strategies for applying guidance methods to multi-modality 3D molecular objects, such as discrete atom types and continuous atom positions.
Although recent studies have begun exploring guidance mechanisms for molecular generation, existing approaches do not fully address this multi-modality. For instance, Luo et al.~\cite{luo2025towards} applied classifier-free guidance to molecule generation on the QM9 dataset~\cite{ramakrishnan2014quantum}, but their approach did not address guidance for discrete molecular attributes such as atom types. By contrast, Nisonoff et al.~\cite{nisonoff2025} developed discrete guidance methods for SMILES \cite{weininger1988smiles} generation, but SMILES representations lack the atom coordinate information needed for 3D conformer generation. 
These gaps motivates the need for a comprehensive evaluation of guidance methods that can handle both continuous and discrete molecular modalities within a unified framework.

This gap motivates our development of a unified framework---MolGuidance that systematically handles both continuous and discrete molecular modalities. 
Morover, building upon Nisonoff's theoretical foundation~\cite{nisonoff2025}, we conduct the first comprehensive investigation of discrete guidance formats, revealing fundamental differences between various mathematical formulations and their practical implications for molecular property control.

In a first kind of implementation, we adopted advanced property guidance methods in the context of \textit{de novo} molecular generation based on the foundation of our flow matching model PropMolFlow (PMF) \cite{zeng2025propmolflow}. We aim to evaluate and benchmark four methods for conditional generation---including the vanilla conditional generation, CFG, AG, and MG---focusing on their ability to generate molecules that meet target property profiles, exhibit high structural validity and diversity, and require less computational overhead. Through this comparative study, we demonstrate the strengths and limitations of each guidance method while simultaneously conducting the first systematic investigation of discrete guidance formats within the CFG framework. Our contributions include discovering that empirically-motivated linear guidance on probabilities outperforms theoretically-derived logarithmic guidance on rate matrices due to stability considerations, and revealing that discrete molecular features (atom types, charges, bonds) exert greater control over molecular properties of generated molecules than continuous geometric arrangements (atomic positions) despite the latter receiving dominant training emphasis. We introduce hybrid guidance strategies that separately optimize continuous and discrete molecular modalities through Bayesian optimization, achieving state-of-the-art performance across guidance methods. Our comprehensive evaluation demonstrates that CFG delivers superior property alignment, AG provides optimal performance balance, while MG struggles in multi-modal molecular settings despite its success in computer vision tasks. We establish benchmarks across both QM9 (5 elements) and QMe14S \cite{yuan2025qme14s} (14 elements) datasets with DFT validation, demonstrating robustness and chemical relevance across different guidance methods and different datasets.


\section*{Results\label{sec:results}}
\newcommand{\polarizability}{$\alpha$}
\newcommand{\gap}{$\Delta\epsilon$}
\newcommand{\homo}{$\epsilon_{\mathrm{HOMO}}$}
\newcommand{\lumo}{$\epsilon_{\mathrm{LUMO}}$}
\newcommand{\dipolemoment}{$\mu$}
\newcommand{\Cv}{$C_{v}$}

\subsection*{Overview of MolGuidance}

\begin{figure}
    \centering
    \includegraphics[width=1.0\linewidth]{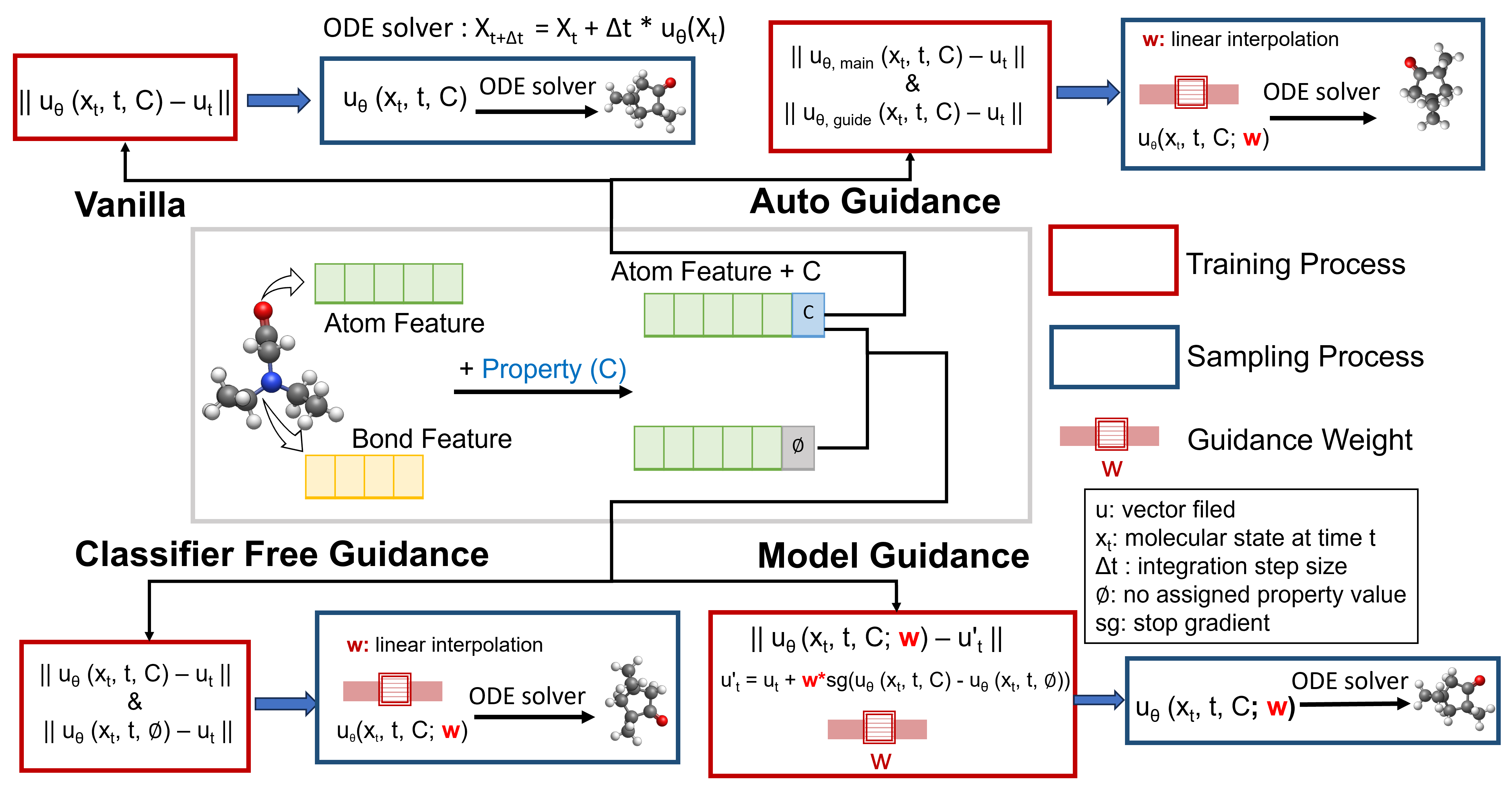}
    \caption{Overview of guidance methods for flow matching-based molecular generation. }
    \label{fig:molguidance_overview}
\end{figure}

\begin{figure}
    \centering
    \includegraphics[width=0.8\linewidth]{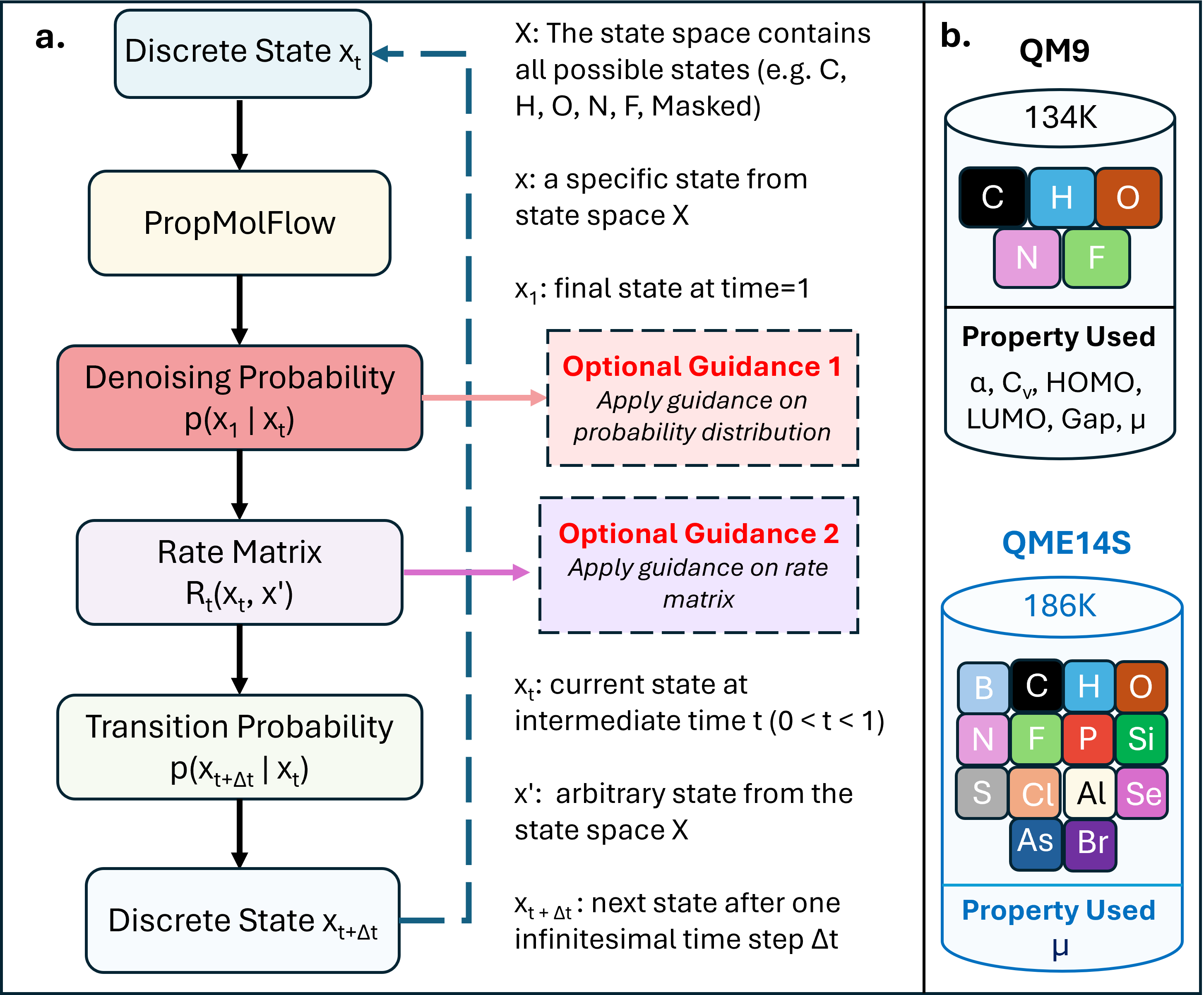}
    \caption{\textbf{a}. Sampling protocol for discrete molecular attributes (atoms, charges, and bonds) in the continuous-time Markov chain (CTMC) process, showing guidance can be applied either to rate matrices or probability distributions. \textbf{b}. Two benchmark datasets—QM9 and QMe14S—used to evaluate different guidance methods, differing primarily in number of elements, dataset size, and target properties.}
    \label{fig:discrete_guidance_and_datasets}
\end{figure}

Flow matching has demonstrated notable success in unconditional molecule generation \cite{hassan2024flow, dunn2024-2}, establishing itself as a powerful framework for learning complex molecular distributions and generating chemically valid structures. However, the integration of advanced guidance methods for conditional generation within flow matching frameworks remains largely unexplored, particularly for property-guided molecular design.
We systematically evaluate three distinct guidance strategies---CFG, AG, and MG---integrated within our previous conditional flow matching architecture PMF.
As illustrated in Figure \ref{fig:molguidance_overview}, our baseline vanilla model learns to transform Gaussian noise into molecular structures conditioned on properties through a flow-based generative process.

The core principle underlying guidance methods is to enhance conditional generation by steering the model toward property-specific molecular distributions, concentrating sampling in targeted regions of chemical space rather than sampling broadly from the unconditional distribution.
Each guidance approach achieves this through fundamentally different mechanisms (Figure \ref{fig:molguidance_overview}): Classifier-free guidance trains both conditional and unconditional models simultaneously, enabling guided sampling through weighted interpolation between conditional and unconditional predictions. The guidance weight (\textit{w}) acts as a tunable parameter controlling the strength of property adherence during inference. Autoguidance employs a deliberately undertrained "inferior model"---with reduced network capacity and/or training epoch---that provides corrective signals to guide the vanilla model toward the target conditional distribution during sampling. In contrast, model guidance integrates guidance directly into the training objective through dynamic interaction between guidance signals and the learning process, eliminating the need for separate guidance computations during inference. 
The details of each guidance method are further discussed in the~\nameref{sec:methodology} Sections.

While guidance methods can be readily applied to the velocity fields of continuous data by linear interpolation (as commonly done in image generation and molecular coordinate generation), certain molecular modalities---such as atom types, charges, and bond order---requires its own approach to respect their discrete nature. 
State-of-the-art methods employ continuous-time Markov chain (CTMC) processes~\cite{campbell2024} to effectively model discrete variables. 
In this framework, illustrated in Figure \ref{fig:discrete_guidance_and_datasets}a, the generative dynamics are governed by a rate matrix~\cite{nisonoff2025}, which specifies the instantaneous transition rates between discrete states. 
At each time step t of the sampling process, the current noisy state $x{_t}$ is passed through the PropMolFlow model to predict the denoising probability distribution $p(x_1|x_t)$, which reflects the model's inferred distribution over the final, clean molecular state. This probability is then used to compute the rate matrix $R{_t}$, from which the transition probability $p(x_{t+\Delta t}|x_t)$ is derived for sampling the next state. 
This two-step process for handling discrete variables offers two distinct strategies for applying guidance ('Optional Guidance 1' and 'Optional Guidance 2'). 
The first approach applies guidance directly to the rate matrix by computing separate conditional and unconditional rate matrices and interpolating between them, thereby blending the dynamics of the two models. This method is theoretically grounded and follows the formulation proposed by Nisonoff et al.~\cite{nisonoff2025}.
In contrast, the second, more empirical approach applies guidance to the denoising probability distribution prior to deriving the rate matrix. Specifically, it interpolates the conditional and unconditional probability predictions to obtain a guided distribution, from which a single, final rate matrix is subsequently computed. 
Furthermore, both approaches support either linear or logarithmic interpolation for combining the conditional and unconditional components. The linear scheme performs a direct, additive blend in probability space, while the logarithmic scheme operates in log-probability space to create a multiplicative blend of the guidance components. (Details are in \ref{sec:math_guidance_formats} of Supporting Information).
These options, which provide multiple pathways for discrete guidance, will be discussed in detail in the following section. 
This flexibility in guidance application---spanning both the stage of intervention (rate matrix vs. probability distributions) and the transformation format (linear vs. logarithmic)---
creates a rich design space for controlling discrete molecular generation processes, as systematically investigated in the following sections.

In this work, we employed two molecular datasets of increasing complexity to evaluate our model (Figure \ref{fig:discrete_guidance_and_datasets}b). The first is the corrected rQM9 SDF dataset available at~\cite{zenodo_propmolflow}, which contains 134K molecules with explicit bond orders and atomic charges across five elements (C, H, O, N, F). We generated molecules by conditioning on six quantum-mechanical properties, including polarizability (\polarizability), heat capacity (\Cv), \homo, \lumo, \gap, and dipole moment ($\mu$),
where the detailed definitions are provided in \ref{sec:more_details} of Supporting Information. The second dataset, QMe14S~\cite{yuan2025qme14s}, comprises 186K molecules spanning 14 elements and we focused on dipole moment as the representative property.
Details of the data correction to obtain the revised rQM9 SDF data are provided in the supplementary information of our previous work~\cite{zeng2025propmolflow}.
Data splitting details are included in the~\nameref{sec:methodology} Section.

Before comparing different property-guided methods, we first systematically investigated optimal strategies for applying guidance to discrete and continuous variables. For discrete variables, we examined two key design choices---applying guidance in the probability-distribution space or directly in the rate-matrix space---and evaluated both linear and logarithmic interpolation schemes to identify the most stable and effective formulation. We then analyzed the guidance mechanism across both discrete and continuous modalities to determine which domain exerts greater influence on target property control. Building on these insights, we implemented a hybrid guidance strategy and employed Bayesian optimization to identify the optimal guidance weights for each method. Finally, using the established optimal protocol, we comprehensively benchmarked and compared four approaches--Vanilla, CFG, AG, and MG--for property-guided molecular generation, evaluating their performance in terms of property alignment, strutural validity, and chemical diversity.

\subsection*{Engineering Guidance Formats for Stable Generation}\label{sec:guidance_domain} 

\begin{figure}
    \centering
    \includegraphics[width=1.0\linewidth]{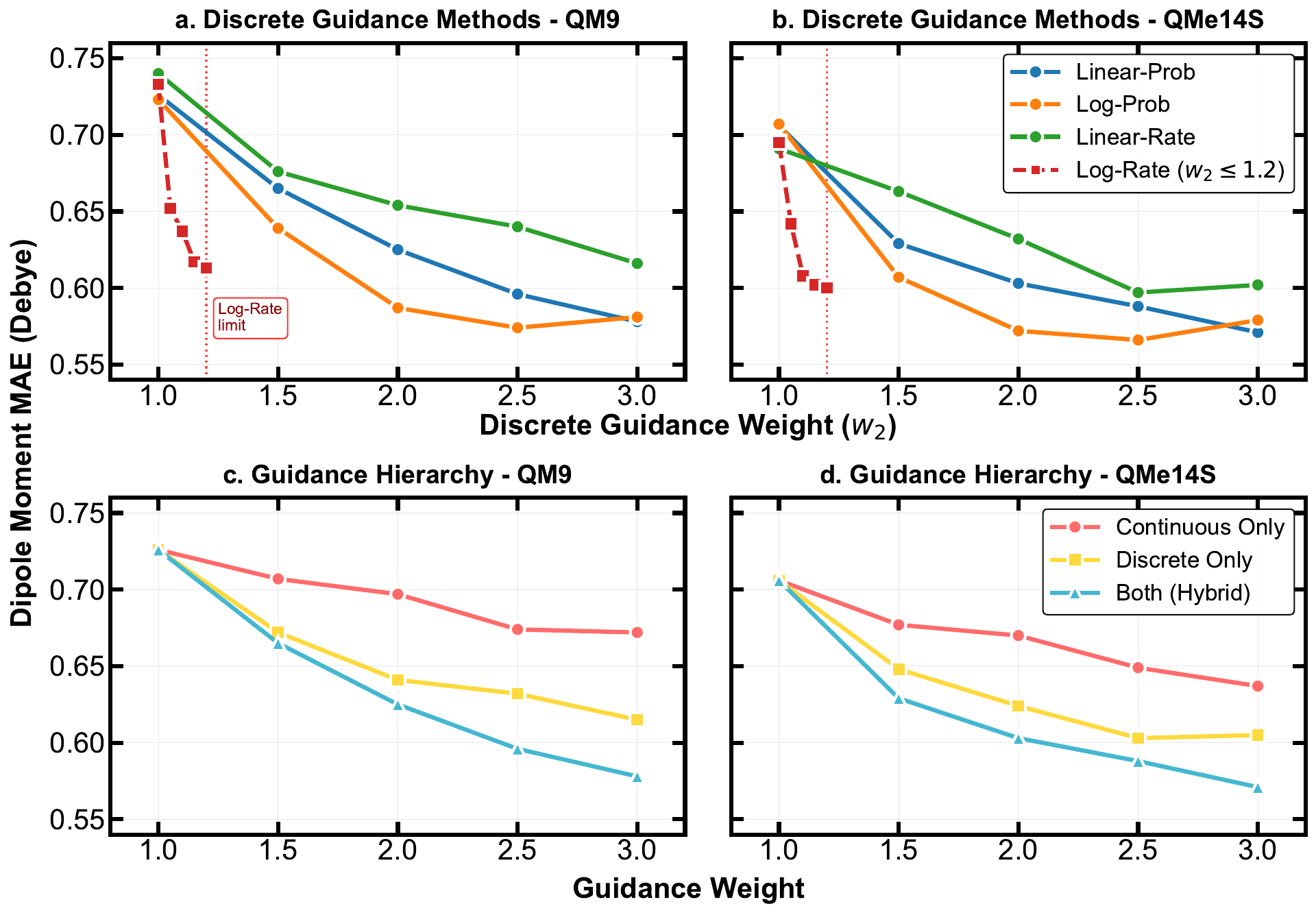}
    \caption{Comparison of guidance methods for dipole moment ($\mu$) prediction across QM9 and QMe14S molecular datasets. Performance is measured as MAE in Debye units as a function of guidance weight. (a-b) Discrete guidance format comparison: Four methods are evaluated—linear guidance on probability (Linear-Prob), logarithm guidance on probability (Log-Prob), linear guidance on rate matrix (Linear-Rate), and logarithm guidance on rate matrix (Log-Rate). The Log-Rate method is constrained to $w_2$ $\leq$ 1.2 because molecular validities drop rapidly at higher guidance weights, indicated by the vertical dotted line. (c-d) Guidance hierarchy comparison: Three strategies are compared—continuous-only guidance (atomic positions), discrete-only guidance (atom types, charges, bonds), and hybrid guidance (both domains). All methods demonstrate improved performance with increasing guidance weight, with optimal values typically occurring between w = 2.0 and 3.0 for numerically stable methods. Lower MAE values indicate better performance and alignment between target and generated molecule properties.}
    \label{fig:guidance_format}
\end{figure}

We first explored the optimal format for applying guidance within our hybrid framework. Our strategy employs separate guidance weights for the continuous and discrete molecular modalities.
In the following equations, we use CFG as the illustrative guidance method, though the formulation generalizes naturally to other guidance schemes.
For the continuous modality (atomic positions), we apply a standard linear interpolation to the velocity fields ($u$), governed by the guidance weight $w_1$:
\begin{equation}
u_{\mathrm{guided}} = (1-w_1)\,u_{\mathrm{unconditional}} + w_1\,u_{\mathrm{conditional}},
\label{eq:linear_prob}
\end{equation}
For the discrete modalities (atom types, charges, and bonds), we systematically investigate four distinct guidance formats, each governed by a separate guidance weight $w_2$. These formats, which represent linear and logarithmic interpolation applied to either denoising probabilities $p_{1|t}$ or rate matrices $R_t(x_t, x')$, are defined as follows:

\begin{equation}
p_{\mathrm{guided}} = (1-w_2)\,p_{\mathrm{unconditional}} + w_2\,p_{\mathrm{conditional}},
\label{eq:linear_prob}
\end{equation}
\begin{equation}
p_{\mathrm{guided}} = \exp\!\big[(1-w_2)\log p_{\mathrm{unconditional}} + w_2\log p_{\mathrm{conditional}}\big],
\label{eq:log_prob}
\end{equation}
\begin{equation}
R_{\mathrm{guided}} = (1-w_2)\,R_{\mathrm{unconditional}} + w_2\,R_{\mathrm{conditional}},
\label{eq:linear_rate}
\end{equation}
\begin{equation}
R_{\mathrm{guided}} = \exp\!\big[(1-w_2)\log R_{\mathrm{unconditional}} + w_2\log R_{\mathrm{conditional}}\big],
\label{eq:log_rate}
\end{equation}

Figure~\ref{fig:guidance_format}a and \ref{fig:guidance_format}b presents representative results for $\mu$ on both QM9 and QMe14S datasets, revealing substantial differences in guidance behavior across discrete weight ranges. 
In these experiments, we systematically varied both the continuous weight ($w_1$) and discrete weight ($w_2$) from 1.0 to 3.0 in increments of 0.5 for three of the four methods. The fourth method, logarithmic guidance on rate matrices, exhibited severe performance drop in molecular structural validity that constrained its discrete weight ($w_2$) exploration to a narrow range of 1.0 to 1.2 with increments of 0.05. 
The y-axis in  Figure~\ref{fig:guidance_format} represents the mean absolute error (MAE) between target $\mu$ values used for conditional generation and the $\mu$ values predicted by a pretrained regressor on the generated molecules, where lower values indicate better property alignment. 

The logarithmic format on rate matrices, while theoretically derived by Nisonoff's work \cite{nisonoff2025}, exhibits severe instability at discrete guidance weights beyond $w_2 = 1.2$, fundamentally limiting its practical applicability. In contrast, the logarithmic guidance on probabilities find its minimum property MAE around $w_2 = 2.5$ as shown in Figure~\ref{fig:guidance_format}a and \ref{fig:guidance_format}b, achieving the best property alignment while maintaining numerical stability across the full weight range. Linear guidance formats show more gradual improvements with increasing weights but require higher guidance strengths to achieve comparable performance. We focus this analysis on CFG because AG shows low molecular structural validity with discrete guidance weights exceeding 2.0, while guidance-aware MG employs a guidance weight embedding, hence lacking in explicit flexibility for tuning guidance weights during sampling, making equivalent comparison infeasible. These findings highlight that empirically-motivated guidance formulations sometimes outperform theoretically-derived approaches in practical molecular generation, suggesting that numerical stability and guidance flexibility are critical considerations beyond mathematical elegance. 
The same trends are also observed on the other five properties of the QM9 dataset (Figure~\ref{fig:s1}).    

\subsection*{Hybrid Guidance is Better than Unimodal  Guidance}\label{sec:hybridg_guidance}

In PropMolFlow, the flow matching framework, the hierarchical loss weights were assigned during training, with atomic position receiving the highest weight, followed by bond order, formal charge, and atom type (see Supporting Information \ref{sec:loss_and_interpolant} for details). This weighting scheme reflects fundamental chemical principles: the three-dimensional arrangement of atoms serves as the primary determinant of molecular geometry. It governs steric interactions, conformational stability, and ultimately defines the structural scaffold upon which discrete molecular attributes (bond order, formal charge, and atom identity) can be consistently determined~\cite{kearnes2016molecular}. Empirical evidence indicates that accurate learning of atomic coordinates markedly enhances the prediction of other molecular modalities, underscoring the central importance of positional information in 3D molecular reconstruction.~\cite{dunn2024-2}

This chemical hierarchy raises a critical question for hybrid guidance strategies: \textit{Does guidance applied to the continuous space (atomic positions) exert a similarly dominant effect on property alignment compared to guidance on discrete variables?} To investigate this question systematically, we conduct controlled experiments comparing the relative impact of continuous versus discrete guidance weights on molecular property alignment. Given our previous finding that logarithmic formats amplify guidance effects but suffer from instability at higher discrete guidance weights, we employ linear guidance on probabilities for discrete components to ensure fair comparison with the inherently linear continuous guidance formulation.

As shown in Figure~\ref{fig:guidance_format}c and \ref{fig:guidance_format}d, hybrid guidance consistently achieves superior property alignment in both data sets for $\mu$, validating our hypothesis that combining continuous and discrete guidance takes advantage of the full chemical hierarchy. 
However, a counterintuitive finding emerges: discrete-only guidance (set no guidance on atomic positions) substantially outperforms continuous-only guidance (set no guidance on atom type, formal charge and bond order), contradicting the training loss hierarchy where atomic positions receive the dominant weight (see Supporting Information \ref{sec:loss_and_interpolant}).
This apparent inconsistency reveals a fundamental distinction between \textit{training} and \textit{guided sampling} for conditional molecule generation. The training and sampling process of a generative molecule model represents different optimization challenges: during training, atomic positions require dominant emphasis to establish accurate molecular geometries and spatial relationships. However, during sampling, discrete molecular features (atom types, charges, bonds) provide greater leverage for precisely targeting specific property values. 

The same trend has been observed across the other five properties in the QM9 dataset (Figure \ref{fig:s2}). The consistent performance gap between discrete and continuous guidance across increasing weight strengths suggests that once robust geometric representations are established through training, discrete chemical features offer more direct control mechanisms for property-guided generation. This finding has important implications for guidance strategy: optimal guidance with respect to molecular property alignment should prioritize the discrete modalities that provide the greatest control over target properties during sampling, in contrast to the greater emphasis on continuous atomic positions during training.


\subsection*{Bayesian-optimized Hybrid Guidance Pushes the State-of-the-art Performance for Property Alignment}\label{sec:property_alignment}

To identify the optimal guidance weights that offer the best alignment of generated molecules with target properties, we employed Bayesian optimization over the joint guidance weights for atomic positions ($w_1$) and discrete variables ($w_2$). 
The logarithmic probability is used because it demonstrates good convergence and stability with increasing guidance weights.
Implementation details of Bayesian optimization are included in the \hyperref[sec:bayes]{Bayesian Optimization of Guidance Weights} Section.
Using CFG conditioning on $\mu$ as the example (Figure~\ref{fig:bayesian}), optimal performance (MAE=202 meV) occurs at w $(w_1, w_2) =(2.71, 1.91)$, and the property alignment performance depends more on the guidance weights  for discrete modalities ($w_2$) than the continuous atom positions ($w_1$).
Figure~\ref{fig:bayesian} also shows examples results for an AG and MG model.
For AG, we found that using a guide model that is more distinct with the main model yields slightly better property alignment in this study (Table~\ref{tab:ag_bayes_best_guide_model}).
The best guidance weights for CFG, AG and MG, and the variation of property MAEs for Bayesian optimization can be found in Table~\ref{tab:best_ws} and Table~\ref{tab:bayes_mae_ranges}.
More discussion about the Bayesian optimization results can be found in the SI section~\ref{sec:best_ws}.

\begin{figure}[!htbp]
   \centering
   \includegraphics[width=0.95\linewidth]{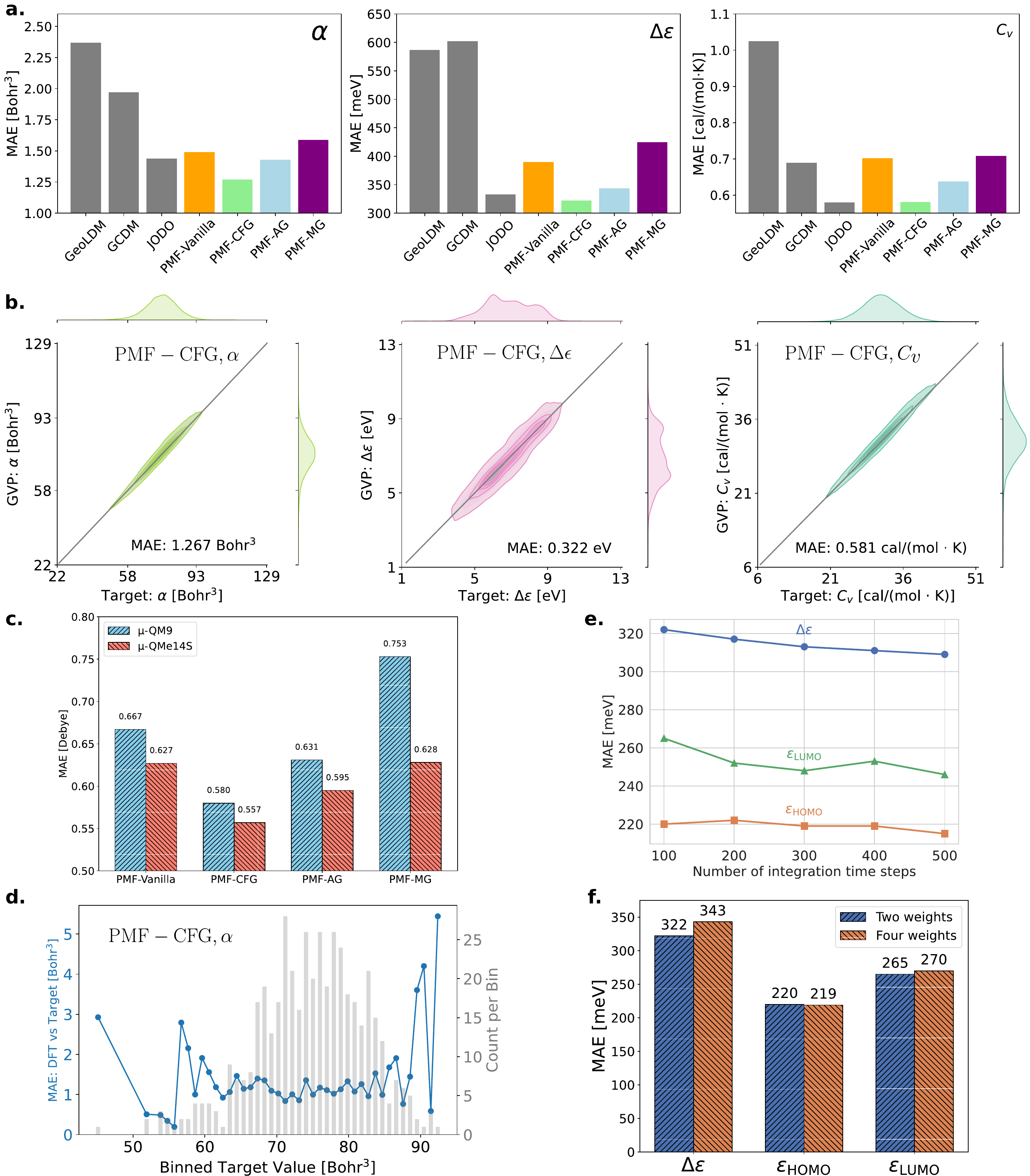}
   \caption{Property alignment performance using different guidance methods. a. Comparison between our guidance methods with three baseline models GeoLDM, GCDM and JODO using property MAEs. Lower numbers indicate better performance. b. One-to-one comparison between target and GVP-predicted property values. c. Performance comparison for models trained on QM9 and QMe14S datasets over dipole moment ($\mu$). d. Mean absolute difference between DFT and Target property values (Left axis) and the target value distribution (Right axis). e. Ablation study on the number of integration steps for PMF-CFG conditioned on $\Delta \epsilon$, $\epsilon_{\mathrm{HOMO}}$ and $\epsilon_{\mathrm{LUMO}}$. f. Property MAEs using Bayesian optimized two guidance weights against those using four guidance weights for PMF-CFG conditioned on $\Delta \epsilon$, $\epsilon_{\mathrm{HOMO}}$ and $\epsilon_{\mathrm{LUMO}}$.}
   \label{fig:property_alignment}
\end{figure}

Our quantitative comparison of conditional generation across different guidance approaches and the three diffusion-based baseline models GeoLDM~\cite{xu2023}, GCDM~\cite{morehead2024}, and JODO~\cite{huang2024} shows that these guidance methods further boost the model's performance. 
Firstly, we analyzed the property alignment performance by comparing PMF guidance methods to three baseline models on polarizability ($\alpha$), HOMO-LUMO gap ($\Delta \epsilon$) and heat capacity ($C_v$)  (Figure~\ref{fig:property_alignment}a). 
We note that the bond order was not used in three baseline models.
Results for the remaining three properties can be found in Table~\ref{tab:property_metric}.
PMF methods outperform previous methods without bond order (GeoLDM and GCDM) by a large margin. 
For example, PMF-CFG achieves a MAE of 1.27 Bohr$^3$ on $\alpha$ compared to 1.97 Bohr$^3$ using the GCDM model.
PMF-CFG achieves the best alignment for four properties--- \polarizability, \gap, \homo~ and \dipolemoment---while remaining competitive on \lumo~ and \Cv~against the state-of-the-art JODO model.
PMF-AG surpasses PMF-Vanilla across all properties, and 
it outperforms JODO on \polarizability, matches its performance on \gap, \homo, \lumo~ and \dipolemoment, and it falls slightly behind on \Cv.
The guidance weight-aware PMF-MG models underperform their vanilla counterparts across all properties, suggesting that jointly learning property constraints and guidance scale embeddings remains challenging for conditional molecule generation.
The quality of property alignment using the CFG guidance is further confirmed by the one-to-one comparison between target and GVP-predicted values (Figure~\ref{fig:property_alignment}b), showing that most points are close to the ideal parity line.

Additionally, we establish new benchmarks for all guidance methods on the more challenging QMe14S dataset using $\mu$ as the target property (Figure~\ref{fig:property_alignment}c). 
Interestingly, the property MAEs on the QMe14S dataset are generally lower than that on QM9 across all guidance methods.
It demonstrates the robustness of our guidance framework across different molecular datasets and chemical space, with all methods maintaining consistent relative performance rankings despite the expanded elemental diversity (5 vs. 14 elements for respective QM9 and QMe14S) and larger molecular size distribution in QMe14S.
Also, the trend of relative differences across guidance are kept for the QMe14S, revealing intrinsic, dataset-independent variation across guidance methods.

Since the property predictor shares the same model architecture with the PropMolFlow generative model, it may exhibit an inductive bias when predicting on molecules generated by PropMolFlow.
To further confirm the performance of property alignment using PMF-CFG, we performed density functional theory (DFT) calculations for 500 molecules selected from the 10k molecules generated at the same level of theory using Gaussian~\cite{g16} as the training data: B3LYP/6-31G(2df,p) for QM9 and B3LYP/TZVP for QMe14S, respectively.
Single-point DFT calculations were conducted on the directly generated molecules for all properties except $C_v$, for which DFT properties of the relaxed molecules were used because of the vibrational frequency issues identified in our prior work~\cite{zeng2025propmolflow}.
DFT results in Figure~\ref{fig:property_alignment}c and Table~\ref{tab:dft_vs_gvp} confirms the property alignment of generated molecules, despite an underestimation of the DFT MAEs against input target property values for $\Delta \epsilon$, $\epsilon_{\rm{HOMO}}$ and $\epsilon_{\rm{LUMO}}$ when a GVP property predictor is used.
To further interpret the MAE distribution across different target values, Figure~\ref{fig:property_alignment}d presents the per-bin MAE for $\alpha$ alongside histograms of the corresponding target-value distribution. 
Larger MAEs tend to appear in regions with fewer data points. 
As a general rule, high prediction errors often arise in sparsely populated regions, such as at both tails of the target-value range.

To explore the limit of the best performing PMF-CFG models, we further increased the number of integration time steps for inference and show the results for  $\Delta \epsilon$, $\epsilon_{\mathrm{HOMO}}$ and $\epsilon_{\mathrm{LUMO}}$. 
Increasing the time steps slightly reduce the property MAEs for these three properties, but it has negligible results for the other three properties (Table~\ref{tab:cfg_n_ts}). 
Moreover, the factorization of flow into the four molecular modalities---atom types, formal charges, bond orders and atom coordinates---allow us to apply guidance separately to each molecular modality.
We also used Bayesian optimization to jointly find the best guidance weights for each of the four, and show comparison between results using four independent guidance weights and using only two guidance weights (that is, one for atom positions and one for all discrete features)  in Figure~\ref{fig:property_alignment}f.
The results show that using four separate guidance weights yields performance comparable to using just two: one for positions and one shared across all discrete variables.
Similar results are observed for AG (Appendix Table~\ref{tab:four_weights_ag}).

\subsection*{Generated Molecules Represent High Structural Validity and Diversity}\label{sec:chemical_validity}
To examine the chemical validity and diversity of generated molecules, we evaluate the molecules using molecule stability, RDKit validity and uniqueness ratio.
Molecule stability evaluates whether the bonding of each atom satisfies simple chemical valency--electron rules.
For example, a zero-charge N atom should have a valency (bond order) of 3 where  an N atom with +1 charge should carry a valency of 4. 
In addition, a molecule is only stable if the total net charge is zero.
RDKit validity examines whether molecules adhere to perception rules so as to ensure consistency of molecular structures within the RDKit framework~\cite{rdkit}.
A molecule is unique if its SMILES is different from others.
\begin{figure}[!htbp]
   \centering
   \includegraphics[width=0.95\linewidth]{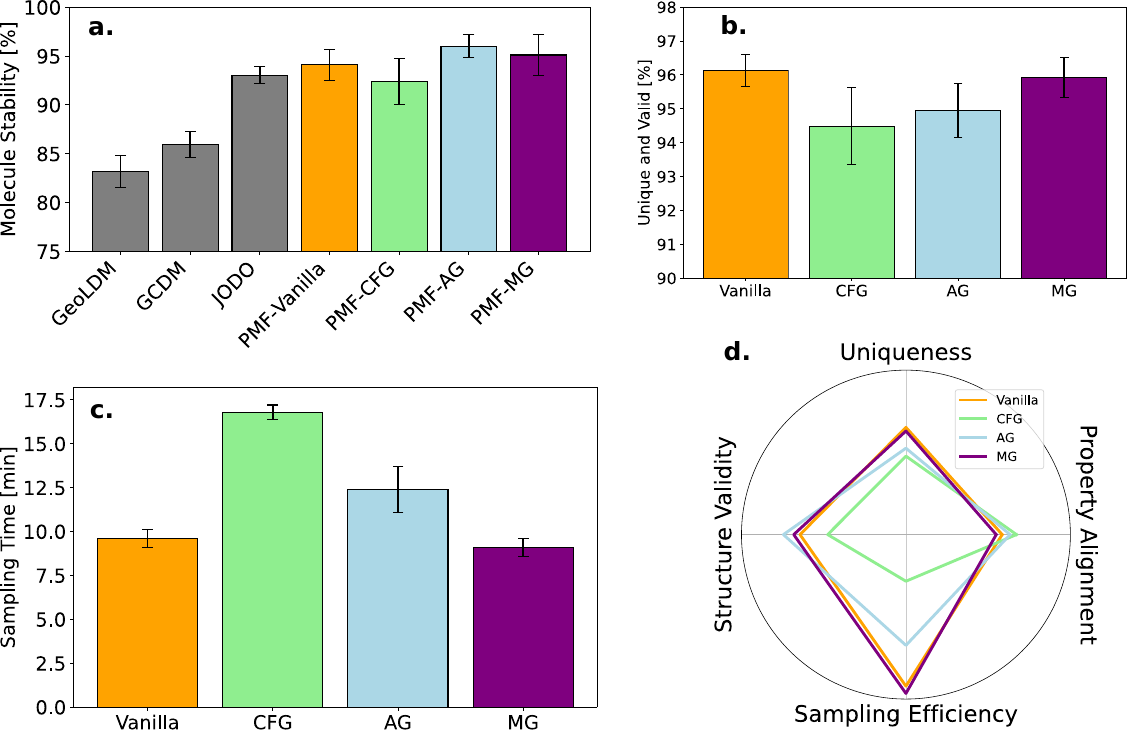}
   \caption{Performance in molecule stability, RDKit validity and uniqueness, sampling efficiency and overall performance for models trained on the rQM9 data. (a) Molecule stability of guidance methods against three baseline conditional models. (b) RDKit validity and uniqueness ratio across our guidance methods. (c) Sampling efficiency across different guidance methods. (d) Overall performance in four dimensions. Results in (a), (b), (c) are reported as the mean $\pm$ std across six molecular properties.}
   \label{fig:other_metrics}
\end{figure}
Results for molecule stability indicate that all guidance methods outperform GeoLDM and GCDM, which omit bond orders, and edge out the previous SOTA, JODO, with explicit bond information in molecular graph representations (Figure~\ref{fig:other_metrics}a).
Although PMG-CFG achieves the best property alignment, it incurs a 2--3.4\% decline in molecule stability compared to the Vanilla conditional model for \gap, \lumo, \dipolemoment, and \Cv; results for \polarizability~and \homo~ remains essentially unchanged. 
In contrast, the PMF-AG models improve molecule stability across all properties relative to PMF-Vanilla, likely because amplifying the differences between main and guide models steer samples away from poorly modeled regions and towards well-learned ones.
Despite their weaker property alignment, MG models surpass their vanilla counterparts in molecule stability, achieving the highest molecule stability for \gap~and \dipolemoment.
The lower molecule stability of CFG, compared to AG and MG, can also be attributed to the higher discrete guidance weights of CFG against those of AG and MG (Table~\ref{tab:best_ws}), and higher guidance weights on discrete modalities correspond to lower molecule stability (Figure~\ref{fig:ablation_ws_molecule_stability}).
Full per-property molecule stability can be found in Table~\ref{tab:property_metric}.
RDKit validity (Table \ref{tab:rdkit_valid}) and PoseBusters validity (Table \ref{tab:pb_valid}) results follow similar trends among AG, CFG, and MG, although the gains against the vanilla conditional generation become less significant. 

To quantify diversity, we calculated the proportion of generated molecules that are both RDKit valid and unique in their SMILES representation. 
The `Valid and Unique' rates for each method are shown in Figure~\ref{fig:other_metrics}b, and per-property results are summarized in Table~\ref{tab:uniqueness}.
Since SMILES has a one-to-one correspondence with 2D molecular graphs, uniqueness serves as a proxy for the quality of guidance and the diversity of generated molecules.
Compared to the vanilla model, CFG exhibits the most notable decline in uniqueness on average, followed by AG and MG.
We also assessed the bond-order entropy (Table~\ref{tab:bond_entropy}), element entropy (Table~\ref{tab:elenment_entropy}), and scaffold diversity (Table~\ref{tab:scaffold_diversity}) of generated molecules as other diversity metrics.
Because each metric evaluates diversity from a single modality or partial view of the molecular structure, the observed trends are not fully consistent.
For instance, CFG  excels in all three metrics in particular for element entropy, while AG also shows high bond-order entropy and element entropy, but falls short in the scaffold diversity.
We also compared the wall-clock time cost for sampling 10000 molecules for each guidance method (Figure~\ref{fig:other_metrics}c). During sampling, both MG and PMF-Vanilla require only a single forward pass, making them the fastest.
CFG and AG perform two forward passes---conditional and unconditional passes for CFG, main's and guide's passes for AG---so they take more time for sampling. 
AG is faster than CFG in sampling because its guide network is much smaller than CFG's unconditional model.

Having examined these individual performance aspects, we now present a holistic comparison to better understand the fundamental trade-offs between guidance methods.
We compared all guidance methods across four dimensions---property alignment, structural validity, uniqueness, and sampling efficiency (Figure~\ref{fig:other_metrics}d). 
For visualization clarity, each metric has been min-max scaled to the range of [0, 1] (see Section~\ref{sec:more_details} for definitions and scaling ranges).
All three guidance strategies outperform the vanilla model in at least one dimension.
The MG approach remains closest to the vanilla model in every dimension, since it applies identical weights to both atomic positions and discrete variables and learning the property embedding and guidance weight embedding jointly can be challenging, which limits its capability to leverage the guidance-weight effects. 
CFG delivers the best property-alignment, outperforming the vanilla models by at least 10 \% relatively across six properties, though at the cost of a modest drop in structural validity and uniqueness. 
AG models also improves property alignment---albeit less than CFG---and improves structure validity over the vanilla baseline.
Both AG and CFG models incur higher computational costs for inference due to two forward passes during sampling.
This comprehensive comparison demonstrates that CFG delivers superior property alignment because the conditional and unconditional model share the same neural network works and the generated molecules is theoretically guaranteed to follow the desired distribution, albeit with trade-offs in structural validity, molecule diversity and sampling efficiency. AG provides optimal performance balancing the four evaluation dimensions, although the method lacks theoretical rigor and the design of the guide model is non-trivial. MG exhibits similar performance as the vanila conditional model, and it struggles to incorporate varied guidance scales, underscoring domain-specific challenges.

\section*{Discussion}
This work presents a comprehensive study of advanced guidance methods for conditional molecule generation. 
Our framework demonstrates robust transferability from QM9 (5 elements) to QMe14S (14 elements) and pushes a new state-of-the-art performance for property alignment and structural validity. 

While our work lays a solid foundation for conditional molecule generation, current study is still focused on a single scaler property. 
Extending the methods to high-dimensional properties such as tensorial properties and to multi-property joint conditions require methodological development and future investigation.
Also, although we studied two strategies to construct the guide model in AG, further exploring the guide models, such as EMA strategy~\cite{karras2024}, may give rise to better performance. 
Besides, guidance based on the linear interpolation of velocity fields may break on general Riemannian folds~\cite{chen2024flow} or with a non-Gaussian base distribution and this line of research represents a promising future direction.
Moreover, recent efforts also focus on inference time optimization and add a control support to steer the generation towards desired directions during sampling~\cite{kim2025inferencetime, liu2025}.

Despite these limitations and challenges, the diverse guidance methods introduced and guidance format discussed  here are readily transferrable to other scientific domains involving multi-modality representations, including \textit{de novo} generation for both inorganic and organic materials and protein design, thereby broadening the impact of robust conditional generation across chemistry, materials science, and beyond.

\phantomsection
\section*{Methods}
\label{sec:methodology}
\newcommand{\p}{\mathbf{p}}
\newcommand{\bigsquarebracket}[1]{\left[#1\right]}
\newcommand{\bigcurlybracket}[1]{\left\{#1\right\}}
\newcommand{\bigbracket}[1]{\left(#1\right)}
\newcommand{\norm}[1]{\lVert #1 \rVert}
\newcommand{\aset}[1]{\{#1\}}
\newcommand{\nprod}{\prod_{i=1}^N}
\newcommand{\nsum}{\sum_{i=1}^N}
\renewcommand{\thefootnote}{\roman{footnote}}

\subsection*{Molecular Representation}\label{sec:mol_rep}

Molecules are represented by fully-connected graphs $G$. 
Graph nodes encode the atomic types $A^i$, charges $C^i$ and positions $X^i$ of a molecule,\footnote{We denote the atom index using a superscript; \textit{e.g.}, position of atom $i$ is denoted $X^i$} and edges are the bond orders between two atoms $E^{ij}$ which are found to enhance structure validity and stability for generated molecules~\cite{vignac2023, dunn2024-2, le2023}. 
Therefore, a molecule can be denoted as $G = (X, A, C, E)$, where  $X = \{X^i\}_{i=1}^N$,  $A = \{A^i\}_{i=1}^N$, $C=\{C^i\}_{i=1}^N$ and $E=\{E^{ij} | i \neq j, i,j \in \{1,2,\cdots, N\}\}$ are the atomic positions, atom types, charges and bond orders, respectively.
Note that atom types, charges and bond orders are discrete categorical variables, while atomic positions are continuous variables.

\subsection*{Joint Flow Matching for Molecule Generation}\label{sec:joint_fm}
Each molecular modality and the molecular graph is learned through a joint flow matching process parameterized by an SE(3) equivariant graph neural network.
Equivariance is critical to improve the model expressivity to describe equivariant properties, as molecules are geometric objects whose properties, such as atomic forces or hyper-polarizability, are subject to equivariance~\cite{satorras2022, batzner2022, thomas2018, xu2025}.
The SE(3) equivariant framework implemented in FlowMol~\cite{dunn2024-1, dunn2024-2} is used as opposed to E(3) frameworks~\cite{hoogeboom2022, xu2023} because breaking the reflection symmetry can generate different molecules due to chirality~\cite{schneuing2024, dumitrescu2024}.
PropMolFlow used here for vanilla conditional generation, is the property-conditioned implementation of FlowMol.~\cite{zeng2025propmolflow}. 
A linear interpolant is used for all molecular modalities.
Vanilla flow matching is adopted for continuous variables, such as atomic positions, whereas discrete modalities, like atom types, bond order, and charges are evolved by discrete flow matching using a Continuous Time Markov Chain (CTMC) process, proposed by~\cite{campbell2024} and~\cite{gat2024}.
Interaction between each molecular modality is achieved by a sequence of node feature, node position, and edge feature updates.
Interpolant and loss function details can be found in Supporting Information \ref{sec:loss_and_interpolant}, and  Model details can be found in Supporting Information \ref{sec:appendix_model}.

\subsection*{Vanilla Conditional Generation\label{sec:vanilla}}
Conditional generation allows users to generate samples aligned with specific requirements. 
This is typically achieved with generative processes parameterized by neural networks that take the target property as an input.
Specifically, we control the outcome by choosing a property and generating a sample from the conditional distribution $p(x_t|c)$ where $c$ is the condition, for example, the property, label or text prompt, and $x_t$ is the noisy data.
In practice, this can be achieved by training a denoiser network $\epsilon_\theta (x_t, t, c)$ for diffusion models or a conditional velocity field $u_\theta (x_t, t, c)$ for a flow matching generative process.
Taking flow matching as an example, the learning objective can be written as:
\begin{equation}\label{eq:vanilla_model}
    \mathcal{L}(\theta) = \mathbb{E}_{t, p_{t|1}(x_t|z, c), p_z} \bigsquarebracket{\norm{u_\theta (x_t, t, c) - u_t}}
\end{equation}
where $u_t$ is the target conditional velocity field at time $t$, and $z$ is a conditioning variable that is normally chosen as $z=(x_0, x_1)$ representing both the initial and final states from a base and target distribution, respectively.
This type of conditional generation is often termed as \textit{Vanilla} conditional generation without explicit guidance.

\subsection*{Classifier-free Guidance}\label{sec:cfg}
In computer vision, vanilla conditional generation trained on complex visual datasets often struggle to reproduce training images due to the finite capacity of neural nets~\cite{karras2024}.
To improve sample quality, \textit{classifier guidance} was introduced by~\cite{dhariwal2021}.
This approach employs an auxiliary classifier $p_\theta(c | x_t)$ to perform low-temperature sampling  by amplifying data points for which the classifier assigns a high likelihood to the target label.  
It approximates a modified distribution:
\begin{equation}\label{eq:modified_prob}
    \tilde{p}_\theta(x_t | c) = p_\theta(x_t | c) \cdot p_\theta(c | x_t)^{w-1}
\end{equation}
Assigning $w>1$ serves as a guidance scale that steers the sampling process toward regions of high classifier confidence.
While effective at increasing alignment with the desired class, classifier guidance requires training an additional classifier on noisy intermediate data, and relies on classifier's gradient, $\nabla_{x_t} \log p_{\theta}(y | x_t)$, to direct samples toward high-likelihood regions, often at the expense of sample diversity.

\textit{Classifier-free guidance} (CFG) is an alternative to classifier guidance with the same effect but does not rely on gradients from a classifier~\cite{ho2022}.
In a CFG approach for flow matching, we train a velocity field $u_\theta(x_t, t, \varnothing)$ without property conditioning and a velocity field with property conditioning $u_{\theta}(x_t, t, c)$.  
During training, a portion of property labels---typically $p_{\rm{uncond}}=0.1$~\cite{ho2022, tang2025}---are dropped and replaced with empty labels to allow both conditional and unconditional objectives to be learned within a single framework.
During sampling, CFG requires two forward passes---one pass with conditioning and another without conditioning---to generate samples, thereby nearly doubling the computational overhead at sampling compared to a vanilla conditional model.
The inference linearly interpolates between these two velocity fields with a weight $w$:
\begin{equation}\label{eq:cfg}
\hat{u}_\theta(x_t, t, c;w) = (1 - w) \cdot u_\theta(x_t, t, \varnothing) + w \cdot u_\theta(x_t, t, c)
\end{equation}
where $w$ is the guidance weight controlling the strength of conditioning. $w=0$ recovers unconditional generation, and $w=1$ corresponds to vanilla conditional generation. 
This linear interpolation of velocity fields is equivalent to linearly interpolating the score functions—the gradients of the log probability density—when the prior is a standard Gaussian~\cite{domingoenrich2025}, which holds for our \textit{de novo} molecule generation setting and results in the modified distribution in Eq.~\ref{eq:modified_prob}.
Values of $w>1$ are often used to further amplify the conditioning signal, which typically leads to stronger adherence to the condition $c$, but also potentially at the cost of even lower sample diversity and validity. 
This reduced validity and diversity have been attributed to the failure of the unconditional model, $u_\theta(x_t, t, \varnothing)$, which faces a more difficult task compared to the conditional model because it only takes a small portion of training budget given by $p_{\mathrm{uncond}}$ while attempting to generate all classes at once~\cite{karras2024}.

\subsection*{Autoguidance}\label{sec:ag}

\textit{Autoguidance} (AG), introduced by~\cite{karras2024}, uses a high-quality main model $D_m$ along with a poor guide model $D_g$ trained on the same task, conditioning, and data splits, but $D_g$  is intentionally degraded, for instance, by having lower model capacity or shorter training. 
This approach is termed \textit{Autoguidance} because it uses a bad version of itself to guide the generation.
The guide model $D_g$ is expected to make similar errors in the same regions as the main model $D_m$, and by subtracting the predictions of the guide model from that of the main model and amplifying the differences by a guidance weight, it pushes the generation away from the weaker model and toward better samples.
In a flow-matching setting, if we denote the main and guide model as $u_m(x_t, t, c)$ and $u_{\rm{g}}(x_t, t, c)$, respectively, the interpolated velocity field at a given weight $w$ reads as:
\begin{equation}\label{eq:mg}
\hat{u}(x_t, t, c;w) = w \cdot u_m(x_t, t, c) + (1-w) \cdot u_{\rm{g}}(x_t, t, c)
\end{equation}

In practice, the main and guide models should be carefully selected to ensure an appropriate quality gap.
The two models should carry similar degradations to remain compatible, while ensuring that the differences are large enough to outweigh random effects such as random initialization of neural networks and random shuffling of training data ~\cite{karras2024ema}.  

\subsection*{Model Guidance}\label{sec:mg}
\textit{Model Guidance} (MG)~\cite{tang2025} offers an alternative to CFG by directly modifying the training objective to include an implicit guidance signal.
Instead of training two separate conditional and unconditional models, MG plugs the guidance weight into the model's training target. 
The model guidance loss then becomes:
\begin{align}\label{eq:mg}
&\mathcal{L}_{\mathrm{MG}}(\theta) =
\mathbb{E}_{t, p_{t|1}(x_t|z, c), p_z} \bigsquarebracket{\norm{u_\theta (x_t, t, c) - \hat{u}_t}} \\
&\hat{u}_t = u_t + w \cdot \mathrm{sg} \bigbracket{\hat{u}_\theta (x_t, t, c) - \hat{u}_\theta (x_t, t, \emptyset) }
\end{align}
where $u_t$ is the ground-truth velocity filed and $\hat{u}_t$ is the modified target. A stopping gradient operation (`sg') is applied to avoid model collapse~\cite{grill2020}. An Exponential Moving Average (EMA) counterpart of the online velocity field, $\hat{u}_\theta (\cdot)$, is normally used to smooth training and provide more stable model predictions.
In addition, the guidance scale/weight $w$ can be fed into neural networks as an additional conditioning input.
The model then learns different guidance weights which offer sampling flexibility to balance sample quality and sample diversity, and it has high sampling efficiency since only one forward for sampling is needed.
How guidance weights are designed for different proportions of data is provided in the Supporting Information~\ref{sec:guidance}.
However, it also creates a difficulty for the models to assimilate the complex interaction between guidance weight embeddings, property embeddings, and molecular graphs.
\begin{table}[h!]
  \centering
  \caption{Comparison of Guidance Approaches}
  \label{tab:guidance_summary}
  \begin{tabular}{c c c c}
    \hline
    \textbf{Method} & \textbf{Extra Model?} & \textbf{Sampling Cost}  & \textbf{Flexible Weights?}\\
    \hline
    Vanilla Conditional Generation & No & 1 forward  & N/A \\
    Classifier‐Free Guidance & No\textsuperscript{$\dagger$} & 2 (cond + uncond) & Yes\\
    Autoguidance & Guide model & 2 (main + guide) & Yes \\
    Model Guidance & No & 1 forward & Optional \\
    \hline
    \multicolumn{3}{l}{\textsuperscript{$\dagger$}\footnotesize Unconditional pass uses same network.}
  \end{tabular}
\end{table}

To summarize, we present the key features of each guidance method in Table \ref{tab:guidance_summary}. 
All methods are implemented within a flow matching generative framework for molecular generation.
While most existing techniques are designed to guide continuous-state generative processes, recent advances have extended guidance to discrete-state spaces~\cite{nisonoff2025}, exploring guidance schemes without special training procedures~\cite{sadat2025}, and investigated how dimensionality impacts the guidance effects~\cite{pavasovic2025}. 

\subsection*{Discrete guidance: Relationship between denoising probability and instantaneous rate matrix entry}\label{sec:log_prob_discrete_guidance}

The work by~\cite{nisonoff2025} proves that in the setting of classifier guidance (or predictor guidance) for a CTMC process. As show in Figure \ref{fig:discrete_guidance_and_datasets}, two key processes require clarification. The first is the conversion from denoising probability to a rate matrix, and the second is the conversion from the rate matrix to transition probabilities.

\paragraph{Denoising Probability to Rate Matrix.} In discrete flow matching with masking processes~\cite{campbell2024}, the neural network learns a denoising model $p_{\theta}(x_1|x_t)$  that predicts the final, clean state ${x_1}$ given the current noisy state ${x_t}$ at time t. This denoising probability will convert to a rate matrix ${R_t}$ to enable sampling via the CTMC framework.

The rate matrix $R_t^\theta(x_t, \cdot)$ used during sampling is obtained as an expectation over the learned denoising distribution~\cite{campbell2024}:
\begin{equation}
R_t^\theta(x_t, \cdot) = \mathbb{E}_{p_{1|t}^\theta(x_1|x_t)} \left[ \bar{R}_t(x_t, \cdot | x_1) \right]
\end{equation}
where $\bar{R}_t(x_t, \cdot | x_1)$ is the data-conditional rate matrix induced by the forward (noising) process $p_{t|1}(x_t|x_1)$.

For the masking process specifically, where $p_{t|1}(x_t|x_1) = \text{Categorical}(t\delta_{x_1,x_t} + (1-t)\delta_{M,x_t})$, and  $\delta$ is the Kronecker delta function. The rate matrix has two components~\cite{campbell2024}: a unmasking entry and a remasking entry.

The rate matrix entry for demasking---from mask token $M$ to actual states---can be written as:
\begin{equation}
R_{t,\text{unmask}}^\theta(x_t, \tilde{x}) = (1 - \delta_{\tilde{x},M}) \cdot p_\theta(x_1 = \tilde{x} | x_t) \cdot \frac{1 + \eta t}{1 - t} \cdot \delta_{x_t,M}
\end{equation}

The rate matrix for remasking reads as:
\begin{equation}
R_{t,\text{remask}}^\theta(x_t, M) = \eta \cdot (1 - \delta_{x_t,M})
\end{equation}
where $\eta$ is the stochasticity parameter that controls the probability of remasking. The total rate matrix is then:
\begin{equation}
R_t^\theta(x_t, \tilde{x}) = R_{t,\text{unmask}}^\theta(x_t, \tilde{x}) + R_{t,\text{remask}}^\theta(x_t, M)
\end{equation}

The diagonal elements $R_t(x_t, x_t)$ are set to ensure conservation of probability flow:
\begin{equation}
R_t^\theta(x_t, x_t) = -\sum_{\tilde{x} \neq x_t} R_t^\theta(x_t, \tilde{x})
\end{equation}

This conversion allows the discrete denoising probability output by the neural network to define the instantaneous transition rates needed for CTMC-based sampling~\cite{campbell2024}.

\paragraph{Rate Matrix to Transition Probability.} 
The probability of jumping from a state $x$ at time $t$ to the next state $\widetilde{x}$ at time $t + \Delta t$ is given by:

\begin{equation}\label{eq:ctmc_cg}
p(x_{t+\Delta t} = \widetilde{x} | x_t = x, y) = \delta_{x, \widetilde{x}} + \frac{p(y| \widetilde{x}, t)}{p(y|x, t)} \cdot R_t(x, \widetilde{x}) \cdot \Delta t + \mathcal{O}(\Delta t^{1+\epsilon})
\end{equation}
where $\Delta t$ defines an infinitesimal time step in the continuous time space. $y$ is the desired property to contion on. $p(y |x_t)$ indicates a predictor/classifier that relates a noisy state sampled at time $t$ to the property, which can be obtained by minimizing a cross-entropy loss. $\delta_{x, \widetilde{x}}$ is the Kronecker function, which can be also denoted as $\delta (x, \widetilde{x})$.
$R_t(x, \widetilde{x})$ indicates the unconditional rate matrix, or it can be written as $R_t(x, \widetilde{x}|\varnothing)$.
$\mathcal{O}(\Delta t^{1+\epsilon})$ is used to denote terms that decay to zero faster when $t \to 0$.

Introducing an inverse guidance temperature $w=1/T$, which is normally termed \textit{guidance strength}, Eq.~\eqref{eq:ctmc_cg} can be converted to:
\begin{equation}\label{eq:ctmc_cg_low_temperature}
p^{(w)}(x_{t+\Delta t} = \widetilde{x} | x_t = x, y) = \delta_{x, \widetilde{x}} + \left[\frac{p(y| \widetilde{x}, t)}{p(y|x, t)} \right]^w\cdot R_t(x, \widetilde{x}) \cdot \Delta t + \mathcal{O}(\Delta t^{1+\epsilon})
\end{equation}
$w>1$ corresponds to a low-temperature sampling, pushing the samples more toward the conditional generation. Then we can also write:
\begin{equation}\label{eq:ctmc_rate_matrix_guidance}
R_t^{(w)}(x, \widetilde{x}|y) = \left[\frac{p(y| \widetilde{x}, t)}{p(y|x, t)} \right]^w \cdot R_t(x, \widetilde{x})
\end{equation}

Reformulating Eq.~\eqref{eq:ctmc_rate_matrix_guidance} using Bayes's theorem, and simplifying it with Taylor expansion (see the Supporting Information in ~\cite{nisonoff2025}), one can obtain the corresponding classifier-free guidance on rate matrix:
\begin{equation}\label{eq:ctmc_cfg}
R_t^{(w)}(x, \widetilde{x}|y) = R_t(x, \widetilde{x}|y)^w \cdot R_t(x, \widetilde{x}|\varnothing)^{1-w}
\end{equation}

Using a linear interpolant, the rate matrix in the formulation of ~\cite{campbell2024} can be also written as: 
\begin{equation}\label{eq:rate_matrix_with_eta}
R_t (x, \widetilde{x}) = \frac{1+\eta t}{1-t} \cdot p^\theta_{1|t}(x_1=\widetilde{x}|x_t=x) \cdot \delta (x, M) + \eta \cdot \bigbracket{1-\delta(x, M)} \cdot \delta(\widetilde{x}, M)
\end{equation}
Eq.~\eqref{eq:rate_matrix_with_eta} holds for both conditional and unconditional generation.
If $\eta=0$, the above equation can be reduced:
\begin{equation}\label{eq:standard_rate_matrix}
R_t (x, \widetilde{x}) = \frac{p^\theta_{1|t}(x_1=\widetilde{x}|x_t=x)}{1-t}  \cdot \delta (x, M) \propto p^\theta_{1|t}
\end{equation}
Hence it can be readily shown:
\begin{equation}\label{eq:guidance_equivalence}
R_t^{(w)}(x, \widetilde{x}|y) = \frac{p^{(w)}_{1|t}(x_1=\widetilde{x}|x_t=x, y)}{1-t}  \cdot \delta (x, M)
\end{equation}
where $R_t^{(w)}(x, \widetilde{x}|y)$ is defined in Eq.~\eqref{eq:ctmc_cfg} and $p^{(w)}_{1|t}(x_1=\widetilde{x}|x_t=x, y)$ refers to the definition in Eq.~\eqref{eq:log_prob} for guidance on discrete variables.

\subsection*{Evaluation metrics} We evaluate guidance methods along four complementary axes.
Property alignment examines how closely the generated molecules' properties match the target properties used during sampling.
This is evaluaed by MAEs of property values of generated molecules compared to the input target values. 
For molecule sampling, each molecule's atom count ($n$) is first drawn from the distribution of QM9 and QMe14S training data, and its property values ($c$) is then conditioned on $n$ to respect the joint distribution on $p(n, c)$.

For structural validity/stability, molecule stability checks the charge--valency consistency. 
Molecular stability is given by proportions of molecules that all atoms are stable and the the net charge of the molecules  are zero if charges are included in the molecular graphs. An atom is stable if it has the correct valency given the formal charge it carries. For instance, a C atom is stable if it has a valency of 4 without charge but a valency of 3 with a -1 charge.
In addition, we also report model performance on RDKit sanitization validity~\cite{rdkit} and PoseBusters validity~\cite{buttenschoen2024}.
RDKit validity corresponds to proportions of molecules that pass the RDkit sanitization.
PoseBusters validity indicates proportions of molecules that pass \textit{de novo} chemical and structural validity tests using the first ten columns of PoseBusters outputs, including sanitization, all atoms connected, valid bond lengths, valid bond angles, no internal steric clashes, and flat aromatic rings.
Structural diversity is evaluated by the `valid and unique' ratios, bond-order entropy, element entropy and scaffold diversity.
`Valid and unique' ratios quantify the number of molecules that can be sanitized in RDKit and are unique in their SMILES representation.
Bond-order entropy is the base-2 Shannon entropy measure of how diverse the bond-type distribution is across a set of generated molecules. Here, four types of bonds---single, double, triple and aromatic bonds---are considered. We first counter the global total of each bond type $n_i$ across generated molecules, and use the counts to generate probability distribution $p_i$ where $p_i = n_i/\sum_i n_i$. The Shannon entropy is then given by: $ H = -  \sum_i p_i \log_2(p_i) $. 
Element entropy is computed analogously to bond-order entropy but measures the diversity of atom type distributions across the generated molecules. For each element type (e.g., C, H, O, N, F for QM9; additionally B, Al, Si, P, S, Cl, As, Se, Br for QMe14S), we count the global total of each element $n_i$ across all generated molecules and compute the Shannon entropy using the same formula. Higher element entropy indicates more diverse elemental compositions.
Scaffold diversity quantifies the structural variety of molecular frameworks using Murcko scaffolds~\cite{bemis1996properties}, which represent the core ring systems and linkers of molecules while removing side chains. We generate the Murcko scaffold for each molecule and compute scaffold diversity as the ratio of unique scaffold structures to the total number of generated molecules. Higher scaffold diversity indicates that the model generates molecules with more varied core structures rather than repeatedly generating molecules based on the same scaffolds.

Computational efficiency  includes training time and sampling efficiency.
Training time is evaluated using the same hardware settings (see more details in \ref{sec:more_details}).  Sampling efficiency is evaluated by the sampling time for 10k molecules and averaged across conditional models for all six molecule properties.
During sampling, PropMolFlow integrates the learned/interpolated velocity fields via Euler's method over 100 evenly spaced time steps.

\paragraph{Scaling factors for metrics in the radar plot.}
Below we elaborate the definition of each metric in the radar plot (Figure \ref{fig:other_metrics}).
If a higher value is preferred (\textit{e.g.}, Bond Diversity, Structure Validity), the metric is transformed via the Eq. \ref{eq:metric_transformation} (left); otherwise the Right equation is used.
\begin{equation}\label{eq:metric_transformation}
x' = \frac{x - x_{\rm{min}}}{x_{\rm{max}} - x_{\rm{min}}} \mbox{and} \qquad x' = \frac{ x_{\rm{max}} - x}{x_{\rm{max}} - x_{\rm{min}}}
\end{equation}
Sampling efficiency uses the min and max scaling factors of 8 and 20 minutes, respectively.
Structure validity is the average of molecule stability and RDkit validity, and the min and max scaling factors are 90\% and 100\%, respectively.
Uniqueness uses the same scaling min and max as that of structure validity.
Property alignment is quantified by the MAEs between the GVP-predicted property values for generated molecules and the input target property values. Lower values are better. The min and max scaling factors are the QM9 lower bound and the QM9 bound given by the \# Atoms shown in Table \ref{tab:property_metric}.

\subsection*{Dataset}\label{sec:dataset}
We evaluate our methods on two molecular datasets of increasing complexity. The first dataset is the QM9 SDF dataset, which provides explicit bond orders and atomic charges~\cite{wu2018moleculenet}.
The original SDF data was found to carry a significant number of charge and bond errors; so instead we use the corrected SDF dataset `rQM9\_v0.sdf' for training.~\cite{zeng2025propmolflow, rqm9}.
After RDKit sanitization, 133k molecules remained and were further partitioned into 100k training, 20k validation and 13k test samples.
The 100k training set was then split evenly following previous work~\cite{hoogeboom2022}: one subset of 50k for training the PropMolFlow molecular generation models and a second subset of 50k for training the property predictor. 
Details of the original QM9 data are provided in Supporting Information \ref{sec:more_details}.

To assess the scalability of our guidance methods on more diverse chemical spaces, we introduced the QMe14S dataset~\cite{yuan2025qme14s}, which contains 186k small organic molecules spanning 14 elements (H, B, C, N, O, F, Al, Si, P, S, Cl, As, Se, and Br). Since this dataset lacks explicit bond and charge information, we employed Open Babel~\cite{o2011open} to infer these molecular attributes. Following RDKit sanitization, 183k molecules remained and were partitioned into 140k training, 25k validation, and 18k test samples. Following the same protocol as QM9, we allocated 70k data for the PropMolFlow model and 70k for training property predictors.

\subsection*{Implementation Detail~\label{sec:experiment_detail}}

\subsubsection*{Property-Conditioned Molecule Generation}\label{sec:prop_cond}

Property-conditioned generation is achieved by the interaction between the node features of a randomly sampled noisy molecular graph and a property embedding.
The property embedding is generated by projecting scalar molecular properties (\textit{e.g.}, dipole moment $\mu$) to a high-dimensional latent space through a shallow multi-layer perceptron (MLP).
Specifically, for AG and CFG, we employ a `concatenate\_sum' operation: the property embedding is concatenated to the node features,  then projected to a latent space via a MLP to match the dimension of node features, followed by a summation operation.
In prior work we have shown that this choice of operation works well across all properties~\cite{zeng2025propmolflow}.
Following the original MG work~\cite{tang2025}, a `sum' operation is used for MG.
Starting from the property-conditioned node feature, the molecular graph is iteratively updated through the joint flow matching process to generate the final molecule.

\subsubsection*{Guidance Implementation}\label{sec:guidance}
We implemented and compared four variants of guidance in the PropMolFlow framework: vanilla conditional generation, classifier-free guidance, autoguidance, and model guidance.

CFG uses $p_{\mathrm{uncondo}}=0.1$,  which denotes the probability of training on unconditional generation during joint training of the conditional and unconditional flow matching models. During sampling, we applied separate guidance weights for continuous atomic positions and discrete variables (atom types, charges and bond orders) to accommodate the hybrid joint flow matching. 

AG uses two types of guide models, $u_{\rm{g}}(x_t, t, c)$, with reduced training time and/or decreased model complexity. 
For under-training, we saved model checkpoints every 20,000 training steps (around every 51 epochs with a batch size of 128). 
For model parameter reduction, we decreased both the hidden dimension size and the number of message-passing layers in the GNN vector field model, reducing the parameter count from 7.1M to 313K.

For MG, we incorporated the guidance weight as an additional input condition, allowing the model to adapt to flexible guidance weights during sampling. We maintained $p_{\mathrm{uncond}}=0.1$ for which guidance weights are set as zeros, and set the proportion of training examples with model guidance to 0.2 and their guidance weights are randomly chosen between 1 and 2, and the remaining data is treated as the vanilla conditional model for which the guidance weights are ones. The model guidance is introduced after 10,000 training steps.
To obtain the modified target velocity field in Eq.~\eqref{eq:mg}, EMA uses a decay rate of 0.9999. 

\subsubsection*{Bayesian Optimization of Guidance Weights}\label{sec:bayes}
The MAEs between target properties and properties of generated molecules are modeled as the optimization objective. 
This Bayesian optimization was performed on top of a Gaussian process via the Scikit-Optimize library~\cite{head2018scikitoptimize}.
We initialized the search with 10 randomly sampled guidance weights to bootstrap the surrogate model, followed by 40 acquisition-function evaluations using the expected improvement (EI) criterion.
To ensure a relatively high structural validity, for AG, $w_1$ and $w_2$ are optimized in the range of $[1, 4.3]$ and $[1, 1.8]$, respectively, whereas for CFG both weights are optimized in the range of $[1, 4]$.
The Bayesian optimization for AG is performed on two guide models, and the guide model with the lowest MAE for each property is reported hereafter.
A scale-aware MG is utilized and the same guidance weights are employed for both atomic positions and discrete variables.
Bayesian optimization for the single MG guidance weight starts with 5 initial points, followed by 10 function evaluations.
For each guidance weight candidate, 1000 molecules were sampled and the objective MAEs were calculated on these molecules.
This procedure was repeated independently for each molecular property.

\subsubsection*{Training details and hyperparamters}\label{sec:hyperparameters}
All guidance models on the QM9 and QMe14S datasets were trained with 8 molecule update blocks.
Atoms contain 256 hidden scalar features and 16 hidden vector features.
Edges contain 128 hidden features.
These models were trained with 2000 epochs with a learning rate of 0.00025 together with an Adam optimizer~\cite{kingma2017} is used for learning the neural networks.
Training and inference for PropMolFlow models used a single NVIDIA A100-SXM4 graphic card with 80GB memory with a batch size of 128.
All models can be trained in about 2--4 days.

\subsubsection*{GVP regressor details}\label{sec:gvp_details}
To be self-consistent, we trained property regressors using Graph Neural Networks based on GVPs.
We appended an MLP layer that takes the final node scalar features as read-out layers to predict the target property.
The parameters of GVP regressors are optimized by minimizing an mean squared error loss.
Separate models were trained for each of the six tested properties. For the QM9 dataset, GVP training uses a disjoint 50k split compared to the other 50k split  used for training the PropMolFlow generative model. For the QMe14S dataset, 70k molecules are used for training the generative model and the other 70k for training the regressor for the property $\mu$.

\section*{Data availability}
The SDF files containing the molecular structures for the QM9 and QMe14S datasets are available at 
\url{https://zenodo.org/records/15700961} and \url{https://zenodo.org/records/16847162}.

\section*{Code availability}
The code to reproduce this work is available at \url{https://github.com/Liu-Group-UF/MolGuidance}.

\section*{Acknowledgment}
The authors are grateful to Philipp Höllmer for bringing the model guidance paper to our attention at a journal club meeting. This work was supported by NSF Grant OAC-2311632 and the AI and Complex Computational Research Award of the University of Florida. S.M acknowledges support from the Simons Center for Computational Physical Chemistry (Simons Foundation grant 839534). The author thank UFIT Research Computing, and the NVIDIA AI Technology Center at UF for computational resources and consultation. 

\bibliographystyle{unsrt} 
\bibliography{molguidance}

\end{document}